\documentclass[10pt,twocolumn,letterpaper]{article}
\usepackage{cvpr}
\usepackage{times}
\usepackage{epsfig}
\usepackage{graphicx}
\usepackage{amsmath}
\usepackage{amssymb}
\usepackage{hyperref}
\usepackage{booktabs,multirow, multicol}
\usepackage{graphicx}
\usepackage{subcaption}
\usepackage{xcolor}
\usepackage{colortbl}

\definecolor{Gray}{rgb}{0.9, 0.9, 0.9}

\begin{document}

\title{\LARGE \bf The Invisible Threat: Evaluating the Vulnerability of Cross-Spectral Face Recognition to Presentation Attacks}

\author{Anjith George and S\'ebastien Marcel \\
Idiap Research Institute \\
Rue Marconi 19, CH - 1920, Martigny, Switzerland \\
{\tt\small  \{anjith.george, sebastien.marcel\}@idiap.ch  }
}

\maketitle
\thispagestyle{empty}

\begin{abstract}

Cross-spectral face recognition systems are designed to enhance the performance of facial recognition systems by enabling cross-modal matching under challenging operational conditions. A particularly relevant application is the matching of near-infrared (NIR) images to visible-spectrum (VIS) images, enabling the verification of individuals by comparing NIR facial captures acquired with VIS reference images. The use of NIR imaging offers several advantages, including greater robustness to illumination variations, better visibility through glasses and glare, and greater resistance to presentation attacks. Despite these claimed benefits, the robustness of NIR-based systems against presentation attacks has not been systematically studied in the literature. In this work, we conduct a comprehensive evaluation into the vulnerability of NIR-VIS cross-spectral face recognition systems to presentation attacks. Our empirical findings indicate that, although these systems exhibit a certain degree of reliability, they remain vulnerable to specific attacks, emphasizing the need for further research in this area.
\end{abstract}

\section{Introduction}

\begin{figure*}[!htbp]
\centering
  \centering
  \includegraphics[width=\linewidth]{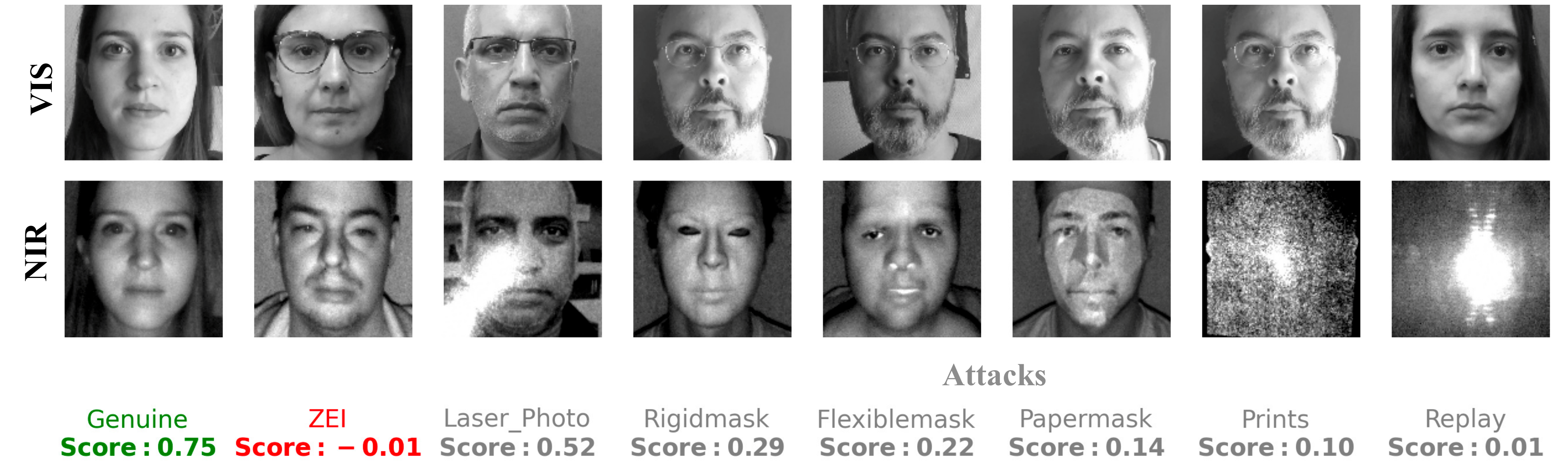}
\caption{Gallery (VIS) and probe (NIR) sample pairs from the WMCA \cite{george2019biometric} VIS-NIR protocol, with match scores (cosine similarity scores normalized to -1 to 1) generated by the SSMB \cite{george2024modality} approach. Examples include genuine pairs, zero-effort impostors (ZEI), and various attack types, with laser photo attacks yielding the highest match scores among attacks.}

\label{fig:samples_attacks}
\end{figure*}

The performance of face recognition systems has improved significantly in recent years \cite{learned2016labeled}, largely due to the availability of large datasets and advances in deep neural network architectures. Face recognition (FR) has become a widely adopted biometric modality due to its ease of use, convenience, and high accuracy. However, its performance can degrade under challenging conditions, such as low-light environments or situations that involve variable illumination. The use of alternative imaging modalities, such as near-infrared (NIR), has emerged as a promising solution to address these limitations.

Cross-spectrum face recognition (CFR) specifically addresses the challenges inherent in visible-spectrum (RGB) imaging by enabling identity verification across different spectral domains. In this framework, individuals enrolled using RGB images can be reliably matched against probe images captured in the NIR domain, even under suboptimal acquisition conditions. Although there is a significant modality gap between RGB and NIR images, recent works in literature have proposed various methods to mitigate this domain discrepancy \cite{tarasiou2024rethinking,hu2024pseudo,nanduri2024semi,guo2020learning}, leading to substantial performance gains on NIR-VIS cross-spectral face recognition benchmarks.

A key benefit of cross-spectrum systems is their capacity to support cross-domain matching without necessitating re-enrollment in the new modality. This feature facilitates the deployment of enhanced, modality-specific sensors in operational environments while maintaining compatibility with legacy RGB-enrolled databases. A comprehensive survey by Anghelone et al. \cite{anghelone2025beyond} further highlights the advantages of cross-spectral face recognition, particularly in applications involving law enforcement, long-range surveillance, and operations conducted under low-light or nighttime conditions.

A critical vulnerability of face recognition systems lies in their vulnerability to presentation attacks (PAs), commonly referred to as spoofing attacks \cite{marcel2014handbook}. These attacks involve the use of artifacts such as printed photographs, replayed videos, or 3D masks to deceive recognition systems, particularly those operating in the visible (VIS) spectrum. Such attacks may be employed to either conceal an individual’s identity or impersonate another subject, known as obfuscation and impersonation attacks. To address this vulnerability, several studies have proposed countermeasures leveraging a variety of detection strategies. Among these, the use of alternative spectral modalities, such as near-infrared (NIR) and thermal imaging has shown promise in enhancing the robustness of face recognition systems to presentation attacks \cite{li2007illumination,george2022comprehensive,george2020can}.

Despite these advancements, the specific vulnerabilities of cross-spectrum face recognition systems, particularly those operating across the NIR and VIS domains, remain underexplored in the prevailing literature. It is to be noted that in CFR setting, only the target modality is available at the time of verification (NIR for instance), and hence models relying on a combination of RGB and NIR are not suitable in this scenario.  While many prior works \cite{bhattacharjee2017you} suggest that NIR-based systems may exhibit inherent resistance to presentation attacks, on the basis that spoof artifacts often manifest differently in the NIR spectrum, this assumption has not been rigorously examined. In this study, we perform a systematic evaluation of the resilience of NIR-VIS cross-spectrum face recognition systems to presentation attacks and provide evidence-based insights and recommendations.

The main contributions of this work are summarized as follows:

\begin{itemize} 
\item We propose a set of new evaluation protocols based on the WMCA dataset \cite{george2019biometric,mostaani2020high} to systematically assess the vulnerability of cross-spectrum face recognition systems to presentation attacks. 
\item We perform a comparative analysis between homogeneous and cross-spectral evaluation protocols, providing insights into the performance. 
\item Through an extensive set of experiments, we demonstrate that cross-spectrum face recognition systems are indeed susceptible to specific presentation attacks, highlighting the need for targeted security enhancements in this domain. 
\item We make all proposed protocols and associated dataset splits publicly available, thereby supporting reproducibility and encouraging further research in cross-spectrum presentation attack detection. 
\end{itemize}

\section{Related work}
\subsection{Cross-spectral Face Recognition}

Heterogeneous Face Recognition (HFR), also referred to as cross-spectral face recognition when the modalities are from different spectra, aims to match facial images captured using different sensing modalities. A main challenge in CFR is the modality gap, i.e., the significant differences in image characteristics between the visible (VIS) spectrum and alternative modalities such as near-infrared (NIR). A range of methods has been proposed in the literature to address this issue. Feature-based approaches, for instance, have shown promise; Klare et al. \cite{klare2010matching} introduced Local Feature-based Discriminant Analysis (LFDA), which combines Scale-Invariant Feature Transform (SIFT) and Multi-Scale Local Binary Pattern (MLBP) descriptors to extract modality-invariant features. Another widely studied direction involves common subspace learning methods, which aim to project images from both source and target modalities into a shared latent feature space, thereby reducing the domain discrepancy and facilitating more effective matching.

Recently, Liu et al. \cite{liu2023modality} proposed a semi-supervised method, Modality-Agnostic Augmented Multi-Collaboration representation for HFR (MAMCO-HFR), leveraging network interactions for discriminative information extraction and introducing a technique for adversarial perturbation-based feature mapping. The work in \cite{sharma2011bypassing} implemented Partial Least Squares (PLS) for linear mapping between modalities. De Freitas et al. \cite{de2018heterogeneous} demonstrated that high-level features from CNNs are domain-independent, employing Domain-Specific Units (DSUs) to minimize domain gaps. Liu et al. \cite{liu2020coupled} introduced techniques such as Coupled Attribute Learning for HFR (CAL-HFR) and Coupled Attribute Guided Triplet Loss (CAGTL) for mapping to a shared space without manual labels. In \cite{george2024heterogeneous}, the authors showed that the lower layers of a network can be made modality-invariant through a teacher-student training approach. Building on this idea, \cite{george2024modality} introduced the SSMB module, which adaptively routes samples in a way that enables the use of a shared latent space for matching both homogeneous and heterogeneous face image pairs.

Many modern approaches to cross-spectral face recognition (CFR) adopt synthesis-based framework, wherein an image from the target modality is first translated into the visible (VIS) domain, followed by face recognition using standard VIS-based networks. Zhang et al. \cite{zhang2017generative} employed Generative Adversarial Networks (GANs) to synthesize photo-realistic VIS images from thermal inputs, introducing the GAN-based Visible Face Synthesis (GAN-VFS) framework. Similarly, the Dual Variational Generation (DVG-Face) model \cite{fu2021dvg} leverages GANs to generate VIS images from heterogeneous modalities, achieving competitive performance on multiple heterogeneous face recognition benchmarks. Liu et al. \cite{liu2021heterogeneous} proposed the Heterogeneous Face Interpretable Disentangled Representation (HFIDR) framework, which disentangles identity-related latent features to enable effective cross-modality synthesis. Despite their promising results, these synthesis-based methods are computationally intensive due to the complexity of generative models and are susceptible to artifacts or hallucinated features that may negatively impact recognition accuracy.

\subsection{Vulnerability Analysis}

Several studies have attempted to assess the vulnerability of face recognition systems to presentation attacks. In \cite{chingovska2016face}, the authors conducted a comprehensive evaluation of face recognition systems under various spoofing scenarios across multiple modalities, including 2D, 3D, and multi-spectral imaging. Their study examined common attack types such as printed photographs, video replays, and 3D mask attacks emphasizing the substantial security risks these pose, particularly in unsupervised or uncontrolled environments. Their evaluation focused on early-stage face recognition systems that relied on handcrafted features and analyzed performance in both the visible (VIS) and near-infrared (NIR) spectra independently. However, this work predates the widespread adoption of deep learning-based models and does not address vulnerabilities in cross-spectral recognition settings, which are increasingly relevant in modern biometric applications.

In \cite{raghavendra2017vulnerability}, authors evaluated the susceptibility of multispectral face recognition systems to spoofing attacks using printed images. It presents a study using a multispectral camera that captures images across seven spectral bands, demonstrating significant vulnerabilities across these bands when faced with high-quality printed face artifacts. However, they have not evaluated the performance of cross-spectral recognition systems. In \cite{mohammadi2018deeply} authors evaluate the vulnerability of face recognition systems that use deep learning to presentation attacks. They investigated the robustness of several face recognition (FR) methods, including deep neural network (DNN)-based systems, against various types of presentation attacks, using multiple public datasets designed for this purpose. The findings highlight that while DNN-based FR systems offer improved recognition accuracy, their vulnerability to spoof attacks is notably high, consistently exceeding 90\% across different testing scenarios. The study underscores a crucial aspect: as face verification accuracy increases, so does the system's vulnerability to attacks.

In \cite{bhattacharjee2017you}, Bhattacharjee et al. demonstrated that certain types of presentation attacks are relatively easy to detect in the near-infrared (NIR) spectrum. Specifically, they observed that images replayed on conventional electronic displays often do not appear in NIR, and similarly, images printed using standard Inkjet printers are typically invisible under NIR illumination. However, they also noted that the use of NIR-reflective inks can render printed images visible in the NIR domain, potentially enabling more sophisticated attacks. Despite these insights, their study did not explore the effectiveness of such attacks within cross-spectral face recognition (CFR) systems, leaving a gap in the understanding of vulnerabilities in cross-modal face recognition scenarios.

From the discussions above, it is evident that, although the vulnerabilities of face recognition systems have been extensively investigated, there is a notable lack of research specifically addressing the security and robustness of cross-spectral face recognition systems. In this study, we systematically evaluate the vulnerability of cross-spectral face recognition systems against presentation attacks.

\section{Evaluation Framework}
\begin{figure}[htbp]
  \centering
  \includegraphics[width=\linewidth]{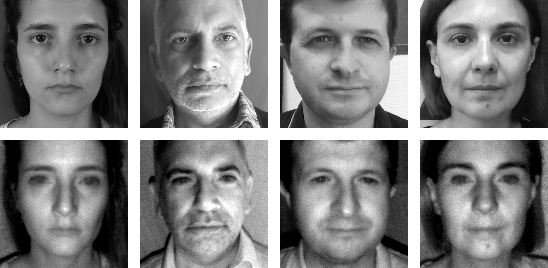}
\caption{VIS and NIR samples from selected identities in the VIS-NIR CFR protocol of WMCA dataset \cite{george2019biometric}. The first row displays images captured in the VIS spectrum, while the second row shows the corresponding NIR images for the same identities.}

\label{fig:samples_pairs}
\end{figure}

\subsection{CFR models}
Cross-spectral Face Recognition (CFR) models enable matching across different imaging modalities. In this work, we specifically address the visible-to-near-infrared (VIS-NIR) matching scenario. Since most CFR datasets are relatively small and insufficient for training models from scratch, models are commonly adapted from pretrained networks trained on RGB data. For our analysis, we select two different CFR systems trained on the MCXFace dataset \cite{george2022prepended,mostaani2020high}, which was specifically designed for heterogeneous (cross-spectral) face recognition. The details of these systems are provided below.

\textbf{Domain Invariant Units (DIU)}: The work in \cite{george2024heterogeneous} introduces a CFR framework named Domain-Invariant Units (DIU) which are trained using a limited amount of paired data within a contrastive framework. This approach leverages a pretrained face recognition model (teacher) to guide the training of a new model (student) to minimize the domain gap and adapt to new variations effectively. The main novelty in DIU involves adapting the lower layers of the student model to learn domain-invariant features while retaining higher-level features trained on extensive RGB datasets. This is achieved by utilizing two loss functions: a cosine contrastive loss to align embeddings from different modalities and a distillation loss to prevent deviation from the teacher model’s embeddings. The method shows superior performance on multiple benchmarks compared to existing state-of-the-art techniques, demonstrating its effectiveness in enhancing pretrained models to handle diverse modalities with minimal amount of paired data.

\textbf{Switch Style Modulation Blocks (SSMB)}: In \cite{george2024modality}, authors introduced a CFR framework that is capable of handling multiple face modalities without requiring explicit knowledge of the target modality during inference. They achieve this though a novel module called  Switch Style Modulation Blocks (SSMB), which automatically route the input images through various domain expert modulators. These modulators adaptively transform the feature maps, significantly reducing the domain gap typically present in cross-modal face recognition tasks. The SSMB allows the system to train end-to-end on a pre-trained face recognition model, transforming it into a modality-agnostic HFR framework. By integrating these blocks into the architecture, the system can dynamically adapt to different input modalities, eliminating the need for modality-specific dataflow paths during inference. The system has been extensively evaluated on HFR benchmark datasets, showing superior performance across diverse conditions and modalities. This versatility is crucial for applications like surveillance where the input can vary significantly where it hard to select the dataflow path accurately, making it a robust solution for real-world scenarios.

\textbf{COTS-FR}: In addition to open-source models, we conducted experiments using a commercial off-the-shelf (COTS) face recognition (FR) software development kit (SDK). The selected system explicitly claims support for both the visible (VIS) and near-infrared (NIR) spectral domains, thereby enabling both homogeneous (within-domain) and heterogeneous (cross-domain) face recognition comparisons.

\section{Experiments}

In this section we detail the process followed for the vulnerability evaluation of the CFR models. The details of the dataset, protocols, and experimental procedure are described in the following subsections.

\textbf{Dataset}: We utilized the WMCA dataset \cite{george2019biometric} for our evaluation since it features a variety of attacks recorded with multiple imaging modalities. The WMCA dataset contains 1941 short video recordings, encompassing both bonafide presentations and presentation attacks from 72 distinct identities. The recordings capture data across several channels, including color, depth, infrared, and thermal. The types of attacks represented in the dataset are diverse, including: (a) paper glasses, glasses with eye designs, printed face images, replayed videos on devices, fake heads, rigid masks such as an Obama plastic Halloween mask,  a transparent plastic mask,  custom-made realistic rigid mask, custom-made realistic flexible mask, and paper masks. While the dataset only contains protocols for presentation attack detection, we newly created protocols to evaluate the vulnerability of cross-spectral face recognition systems. Figure \ref{fig:samples_pairs} shows some samples of bonafides for VIS-NIR protocol.

\textbf{Creation of protocols}: For the vulnerability analysis, we designed new evaluation protocols using a selected subset of attacks from the WMCA dataset. Our focus is specifically on impersonation attacks, which include silicon masks, printed photos, and video replays. The print attacks vary by printing method, using both inkjet and laser printers, and by paper type, with both glossy and matte finishes. The mask attacks include custom silicon masks as well as rigid masks. Replay attacks are conducted using video replays displayed on an iPad.

For our evaluation, we created two separate splits for each protocol: a development (dev) set containing only bona fide samples, and an evaluation (eval) set comprising both bona fide and impersonation attack samples. In both sets, a source modality is used for enrollment and a target modality for probing. The subjects in the dev and eval sets were mutually exclusive.

We introduce two protocols for the evaluation:

\begin{itemize}
    \item VIS-VIS: A homogeneous matching setting used as a baseline to assess changes in vulnerability between standard face recognition and cross-modal settings.
   \item VIS-NIR: The primary cross-spectral protocol of interest, where identities are enrolled using VIS images and probed using NIR images.
\end{itemize}

Figure \ref{fig:samples_attacks} presents examples of bonafide pairs, impostor pairs, and presentation attacks, with gallery images in the VIS domain and probe images in the NIR domain, along with their corresponding match scores as produced by the CFR system.

\textbf{Evaluation}: To perform the evaluation, we first determine the score threshold corresponding to a false match rate (FMR) of 0.1\% on the development set for each protocol. This simulates setting the system threshold to a specific operating point, as would be done in a real-world deployment. For the vulnerability analysis, this threshold is then applied to the evaluation set to compute the final metrics. We denote this threshold as $\tau_{0.1}$, representing the score value at which the FMR equals 0.1\% on the development set.

\subsection{Metrics}

IAPMR, or Impostor Attack Presentation Match Rate \cite{ramachandra2017presentation} \cite{ISO1}, refers to how often a biometric system mistakenly accepts fraudulent presentations as genuine. During verification, users provide a biometric sample and a claimed identity. These fall into three categories: Genuine (both sample and identity are authentic), Zero-effort impostor (ZEI, where the sample is authentic but the identity claimed is not), and Impostor PA (the sample and identity match but neither belong to the actual user). The system's goal is to accept only genuine presentations and reject impostor ones. IAPMR measures the system's vulnerability by quantifying the rate at which impostor PAs are erroneously accepted as genuine.

In our experimental scenario, the threshold for accepting or rejecting an identity in a face recognition system is established at a False Match Rate (FMR) of 0.1\% during the development phase, using a specific protocol's development set. This threshold is then applied to the evaluation set to mirror conditions similar to real-world deployment of a Cross-spectrum Face Recognition (CFR) system. The system's recognition accuracy is assessed using metrics like the FMR and False Non-Match Rate (FNMR). Additionally, the system's threshold is determined using cosine distance calculations based on genuine scores and zero-effort impostor scores. This setup facilitates the evaluation of the system's vulnerability to reject attacks, quantified by the Impostor Attack Presentation Match Rate (IAPMR).

\subsection{Evaluation pipeline}

The WMCA dataset \cite{george2019biometric} includes face landmark annotations extracted using the MTCNN face detector \cite{zhang2016joint}. During preprocessing, all images are aligned and cropped to a standardized resolution of $112 \times 112$. To ensure compatibility with the face recognition (FR) architecture, single-channel modalities are replicated across three channels. For each subject, an embedding is generated using the HFR model for enrollment. Similarity scores are then computed by comparing the reference embedding with probe embeddings using cosine similarity. Score files are produced by evaluating each enrolled subject against all probe samples, including those containing presentation attacks.

\subsection{Results}

\textbf{Face recognition performance}: Before conducting the vulnerability evaluations, it is essential to first assess the face recognition performance of the selected CFR models. Notably, the CFR systems selected in this study are capable of handling both homogeneous (VIS-VIS) and heterogeneous (VIS-NIR) matching without requiring any configuration changes. We evaluate their performance on the development set for both protocols, with the results presented in Table \ref{tab:combined_protocol_cfr}. As shown in the table, both CFR models achieve perfect accuracy in VIS-VIS and VIS-NIR matching, demonstrating strong performance in both homogeneous and cross-modal scenarios.

\begin{table}[h!]
\centering
\caption{Face recognition performance metrics for VIS-VIS (RGB) and VIS-NIR (NIR) protocols.}
\resizebox{0.95\columnwidth}{!}{%
\begin{tabular}{@{}llcccc@{}}
\toprule
\textbf{Protocol} & \textbf{Method}   & \textbf{AUC} & \textbf{EER} & \textbf{VR (FMR=0.1\%)} & \textbf{VR (FMR=1\%)} \\ \midrule
\multirow{2}{*}{VIS-VIS}  
                          & SSMB \cite{george2024modality}     & 100 & 0.0   & 100         & 100         \\
                          & DIU \cite{george2024heterogeneous}      & 100 & 0.0   & 100         & 100         \\ \midrule
\multirow{2}{*}{VIS-NIR}  
                          & SSMB \cite{george2024modality}     & 100 & 0.0   & 100         & 100         \\
                          & DIU \cite{george2024heterogeneous}      & 100 & 0.0   & 100         & 100         \\ \bottomrule
\end{tabular}
\label{tab:combined_protocol_cfr}
}
\end{table}

Additionally, the HFR models were evaluated on seven widely-used face recognition benchmark datasets. These include Labeled Faces in the Wild (LFW) \cite{huang2008labeled}, Cross-Age LFW (CA-LFW) \cite{zheng2017cross}, Cross Pose LFW (CP-LFW) \cite{zheng2018cross}, Celebrities in Frontal-Profile in the Wild (CFP-FP) \cite{sengupta2016frontal}, and AgeDB-30 \cite{moschoglou2017agedb}. We report the accuracy achieved on each dataset. As shown in Table \ref{tab:hfr_fr_perf}, the HFR models demonstrate good performance across these benchmarks despite being specifically trained for a particular modality combination.

\begin{table*}[h]
  \centering
  \caption{Performance of the HFR models on standard face recognition benchmarks. }
  \label{tab:hfr_fr_perf}
  \resizebox{0.75\textwidth}{!}{
\begin{tabular}{llllll}
\toprule

                  \textbf{Model} &           \textbf{LFW} ~\cite{huang2008labeled} &         \textbf{CALFW}~\cite{zheng2017cross} &         \textbf{CPLFW}~\cite{zheng2018cross} &        \textbf{CFP-FP}~\cite{sengupta2016frontal} &      \textbf{AGEDB-30}~\cite{moschoglou2017agedb} \\\midrule

DIU \cite{george2024heterogeneous} & 99.72 ± 0.26 & 95.70 ± 1.02 & 93.32 ± 1.03 & 96.84 ± 1.06 & 97.43 ± 0.84 \\
SSMB \cite{george2024modality}  & 99.78 ± 0.29 & 95.85 ± 1.07 & 91.50 ± 1.30 & 91.81 ± 1.71 & 97.62 ± 0.87 \\
\bottomrule
\end{tabular}
}
\end{table*}

\textbf{Vulnerability Analysis}: For the vulnerability analysis, experiments are performed on the evaluation set using the VIS-NIR protocol. For each protocol, we apply the threshold corresponding to a 0.1\% false match rate (FMR), as determined from the development set, to compute the final evaluation metrics. Although our primary focus is on the VIS-NIR cross-modal face recognition (CFR) setting, we also evaluate the VIS-VIS protocol to provide a comparative baseline against homogeneous face recognition performance.

\begin{figure*}[htbp]
\centering
\begin{subfigure}{.49\linewidth}
  \centering
  \includegraphics[width=\linewidth]{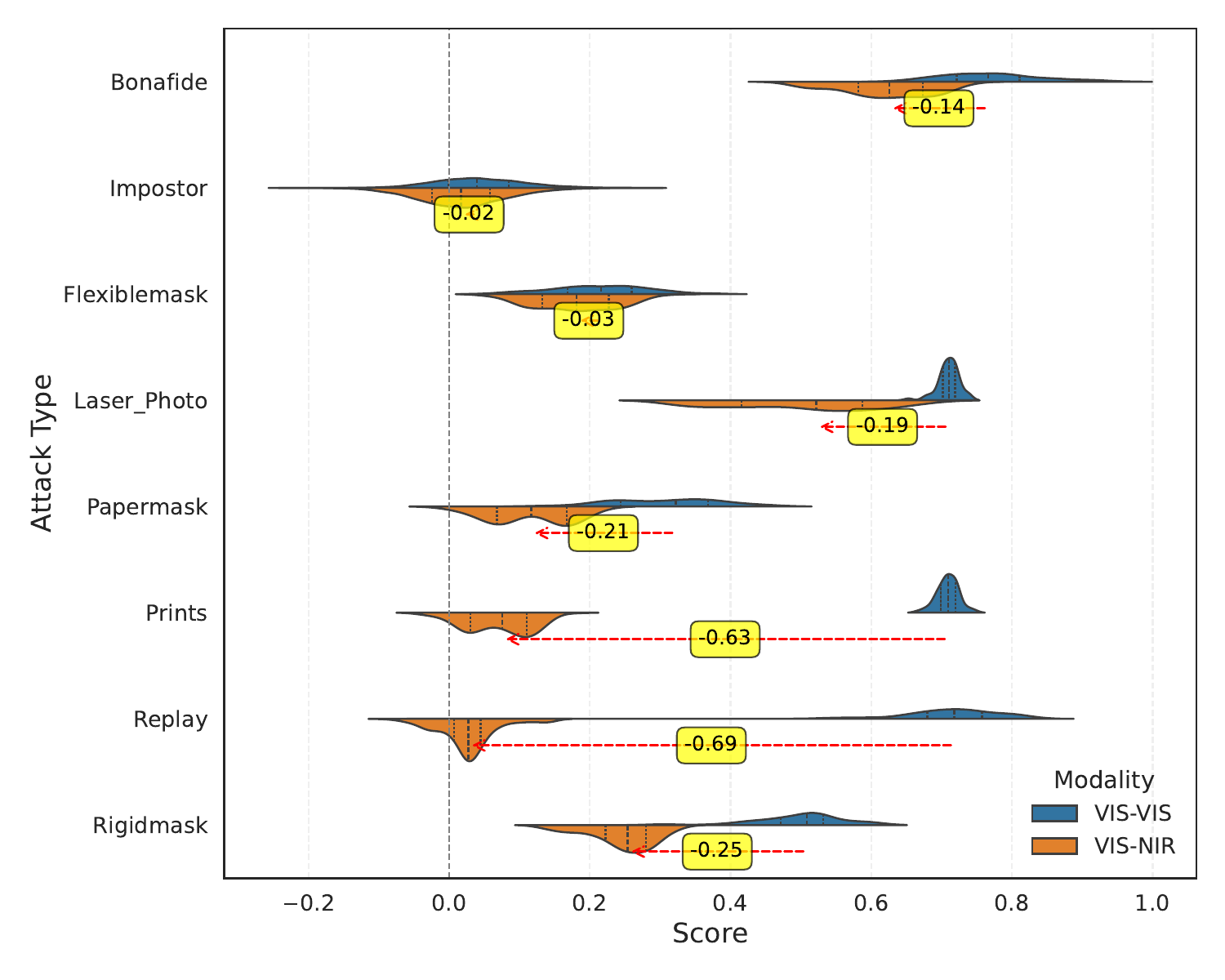}
  \caption{DIU}
\end{subfigure}%
\hfill
\begin{subfigure}{.49\linewidth}
  \centering
  \includegraphics[width=\linewidth]{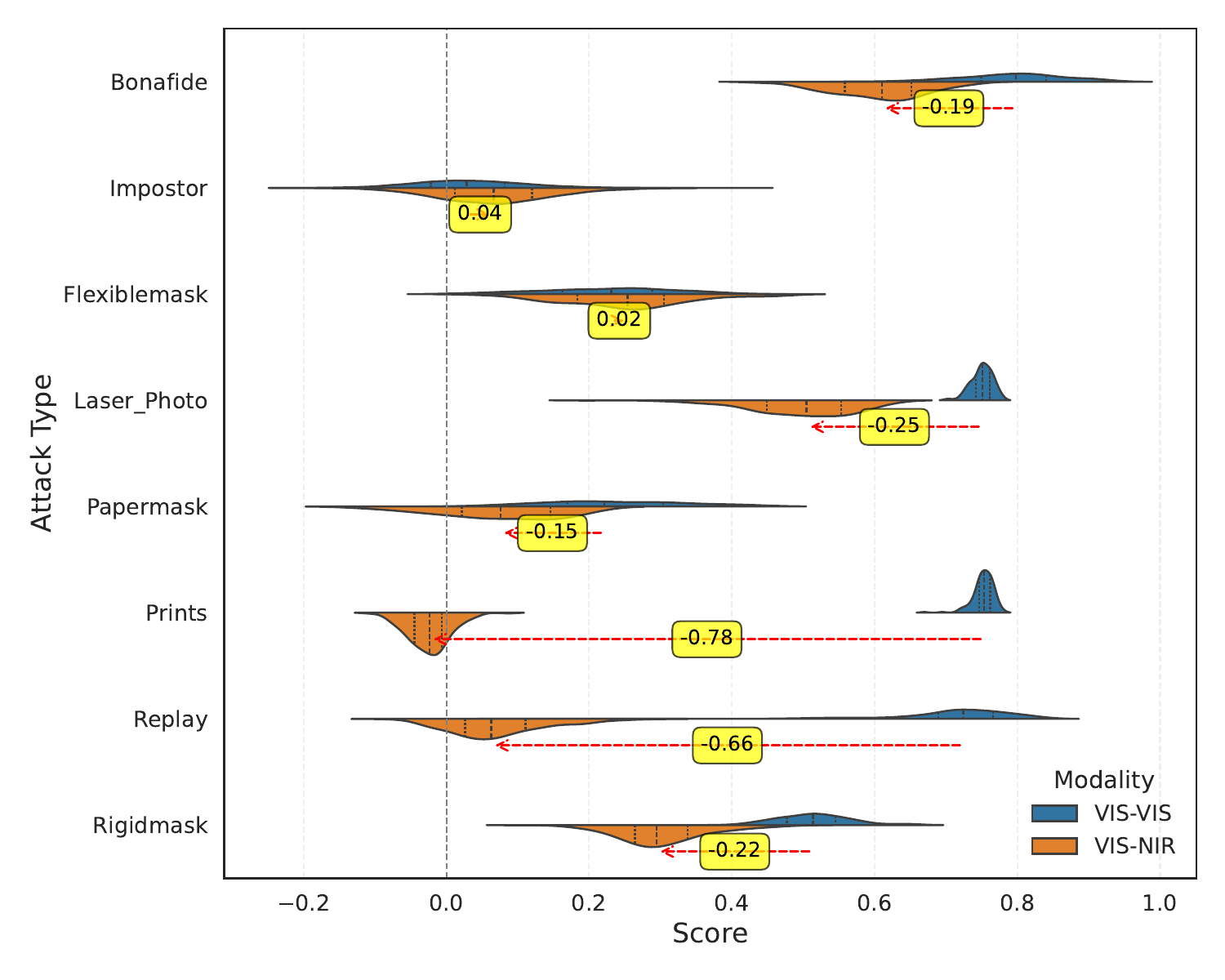}
  \caption{SSMB}
\end{subfigure}%
\hfill

\caption{Score distributions across all categories for both the DIU and SSMB HFR systems are depicted in the plots. Each plot illustrates the score distributions for Bonafide, Impostors (ZEI), and other attack types for a specific HFR system. Higher scores for attacks would indicate increased attack potential. The red arrow indicates the shift in distribution from VIS (blue) to NIR (orange) modalities. A leftward shift (negative) for attacks signifies decreased vulnerability to that specific attack. }

\label{fig:hist_pa_species_diu_ssmb}
\end{figure*}

To better understand the impact of different attacks on the Cross-spectral Face Recognition (CFR) system, we conducted a more fine-grained analysis of the score distributions. Figure \ref{fig:hist_pa_species_diu_ssmb} illustrates the distribution of scores for various attack types under both VIS-VIS and VIS-NIR protocols, providing insights into the differing vulnerabilities of traditional Face Recognition (FR) and CFR systems. The plots reveal shifts in the median scores of each attack category between the VIS-VIS and VIS-NIR scenarios. A rightward shift indicates increased attack effectiveness, while a leftward shift suggests reduced impact.

As observed in the figure, most attacks show a leftward shift in the VIS-NIR setting, implying a generally lower attack potential. This is expected, as many attacks either become invisible or significantly degraded in the near-infrared (NIR) spectrum. For example, replay attack presentations using electronic displays like iPads pose a significant threat in the VIS setting but are largely ineffective in NIR due to the lack of reflectance. Similarly, print attacks using Inkjet printers are effective in VIS but have minimal impact in NIR, as the ink does not reflect well in that spectrum.

Attacks involving masks and wearable disguises also demonstrate reduced effectiveness in NIR compared to VIS, attributed to differences in reflectance characteristics between visible and NIR light. Interestingly, the most effective attack in the VIS-NIR setting is the laser-printed photo attack, as laser prints are highly reflective in the NIR spectrum, making them more challenging for the CFR system to detect.

Table \ref{tab:combined_nir_rgb_iapmr_attacksplit} presents the Impostor Attack Presentation Match Rate (IAPMR) for each CFR system under both VIS-VIS and VIS-NIR protocols. IAPMR quantifies the system's vulnerability by measuring the percentage of attack samples incorrectly accepted as genuine users; thus, higher values indicate greater susceptibility. An ideal system would achieve an IAPMR of 0, signifying perfect discrimination between genuine and attack presentations.

The results show that CFR systems operating in the VIS-NIR setting generally exhibit lower vulnerability compared to their VIS-VIS counterparts. Specifically, the aggregate (ALL) IAPMR drops 84.55\% $\xrightarrow{}$ 15.77\% for SSMB, 85.83\% $\xrightarrow{}$ 32.37\% for DIU, and 80.10\% $\xrightarrow{}$ 22.60\% for COTS-FR when moving from homogeneous to cross-spectral scenarios. In addition to aggregate performance, we also report IAPMR values for the most effective attack types, laser photos and masks. As shown in the table, laser-printed photo attacks are especially potent, with IAPMR values reaching 96.97\%, 100\%, and 100\% across the evaluated systems, underscoring a critical vulnerability. Notably, even commercial off-the-shelf face recognition (COTS-FR) systems are highly susceptible to such attacks.

\begin{table}[h!]
\caption{IAPMR metrics for VIS-NIR and VIS-VIS protocols under various PA species.}
\centering
\resizebox{0.95\columnwidth}{!}{%
\begin{tabular}{@{}llccc@{}}
\toprule
\textbf{Protocol} & \textbf{Metric}             & \textbf{SSMB} \cite{george2024modality}   & \textbf{DIU} \cite{george2024heterogeneous}  & \textbf{COTS-FR}\\  \midrule
\multirow{3}{*}{VIS-NIR} & IAPMR-All attacks   & 15.77  & 32.37  & 22.60\\
                     & IAPMR-Masks         & 15.17  & 58.91  & 33.00 \\
                     & IAPMR-Laser Photos  & \textbf{96.97}  & \textbf{100.00}  & \textbf{100.00 }\\ \midrule
\multirow{3}{*}{VIS-VIS} & IAPMR-All attacks   & 84.55  & 85.83 & 80.10 \\
                     & IAPMR-Masks         & 66.67  & 64.02 & 50.40 \\
                     & IAPMR-Laser Photos  & \textbf{100.00} & \textbf{100.00}  & \textbf{100.00} \\ \bottomrule
\end{tabular}
}
\label{tab:combined_nir_rgb_iapmr_attacksplit}
\end{table}

\begin{figure*}[htbp]
\centering
\begin{subfigure}{.32\textwidth}
  \centering
  \includegraphics[width=\linewidth]{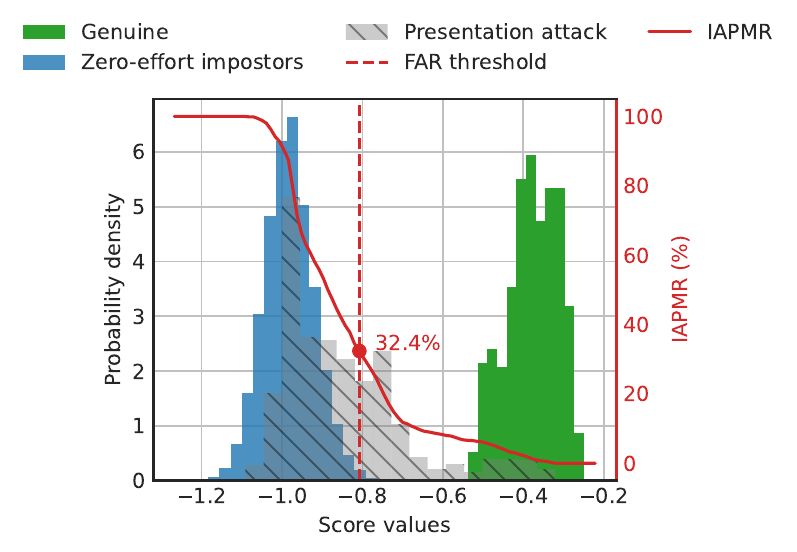}
  \caption{DIU - All PAs}
\end{subfigure}%
\hfill
\begin{subfigure}{.32\textwidth}
  \centering
  \includegraphics[width=\linewidth]{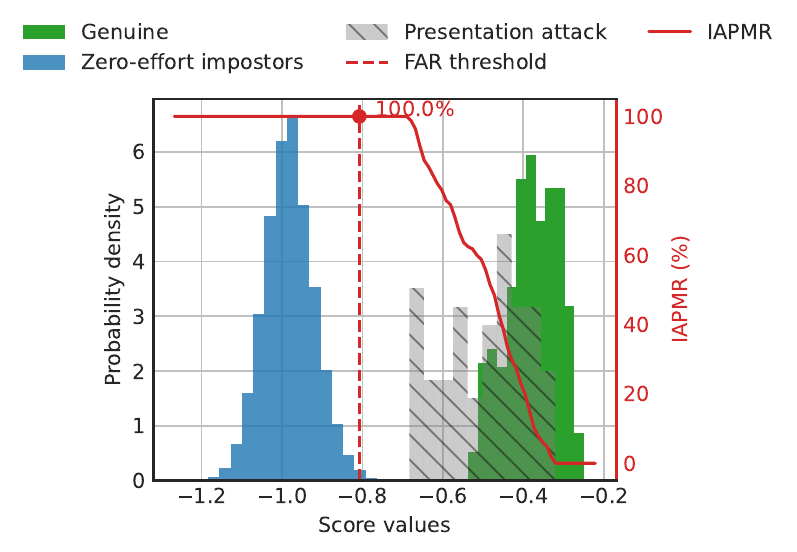}
  \caption{DIU - Laser Photos}
\end{subfigure}%
\hfill
\begin{subfigure}{.32\textwidth}
  \centering
  \includegraphics[width=\linewidth]{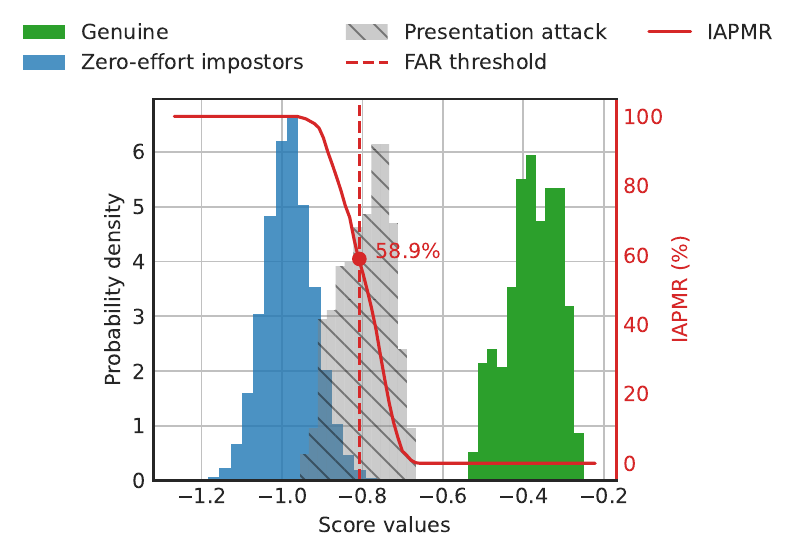}
  \caption{DIU - Masks}
\end{subfigure}

\begin{subfigure}{.32\textwidth}
  \centering
  \includegraphics[width=\linewidth]{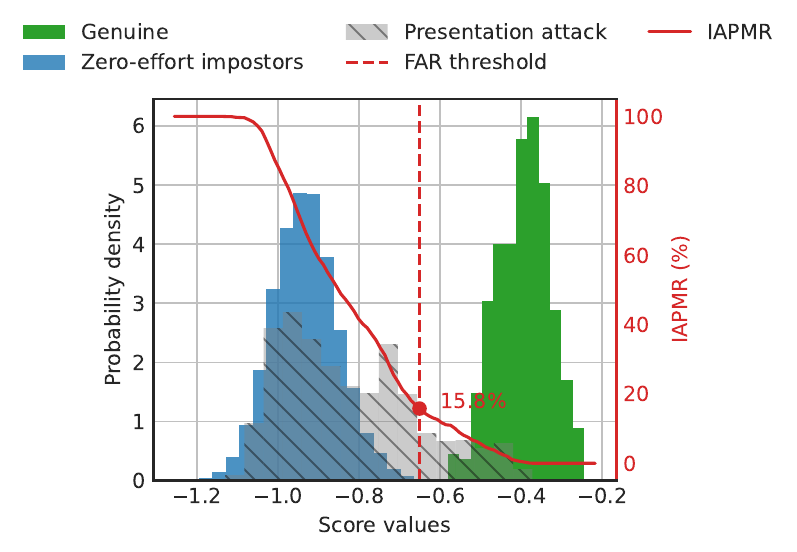}
  \caption{SSMB - All PAs}
\end{subfigure}%
\hfill
\begin{subfigure}{.32\textwidth}
  \centering
  \includegraphics[width=\linewidth]{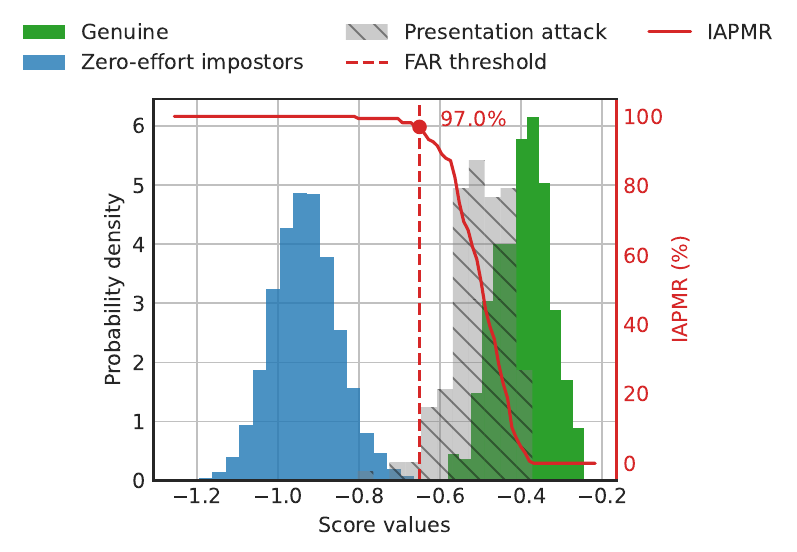}
  \caption{SSMB - Laser Photos}
\end{subfigure}%
\hfill
\begin{subfigure}{.32\textwidth}
  \centering
  \includegraphics[width=\linewidth]{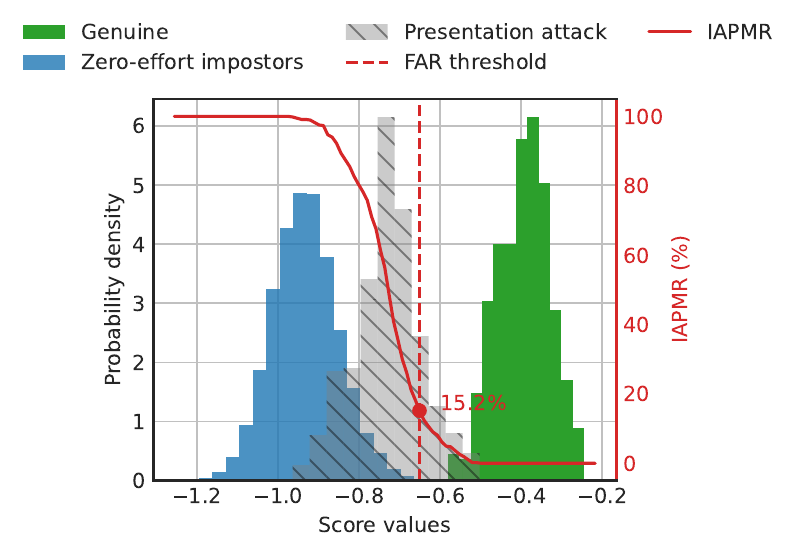}
  \caption{SSMB - Masks}
\end{subfigure}

\caption{Score distributions (VIS-NIR Protocol) for two HFR systems (first row DIU \cite{george2024heterogeneous}, second row SSMB \cite{george2024modality}) with different PA combinations (All PAs, Laser Photos, Masks).
Each plot shows histograms of genuine (green), ZEI (blue), and attack (gray) scores. The red dashed line marks the FMR 0.1\% threshold (from the licit protocol's development group), while the solid red curve represents IAPMR across thresholds. The IAPMR at the given threshold is found at the curve's intersection with the dashed line.}

\label{fig:vulnhist_vis_nir}
\end{figure*}

\begin{figure*}[htbp]
\centering
\begin{subfigure}{.32\textwidth}
  \centering
  \includegraphics[width=\linewidth]{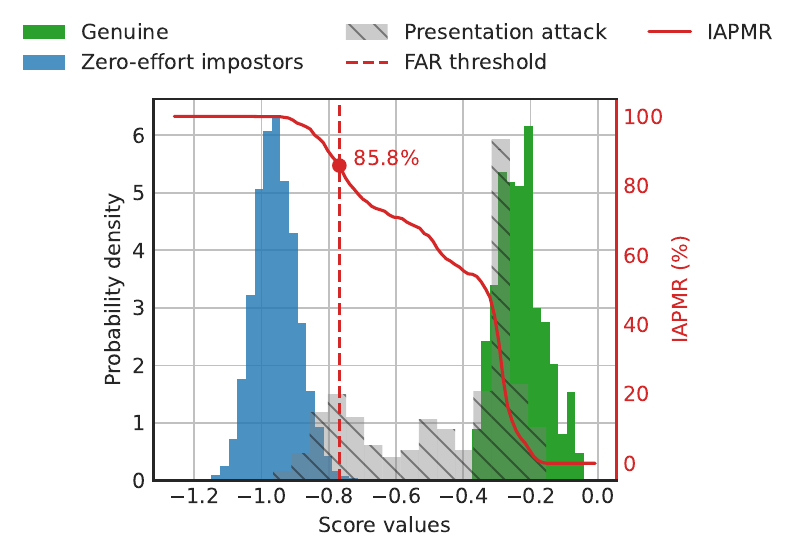}
  \caption{DIU - All PAs}
\end{subfigure}%
\hfill
\begin{subfigure}{.32\textwidth}
  \centering
  \includegraphics[width=\linewidth]{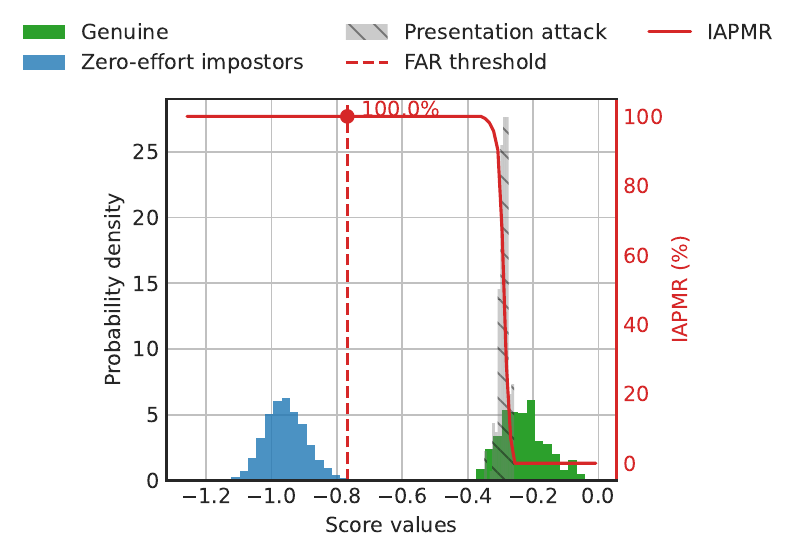}
  \caption{DIU - Laser Photos}
\end{subfigure}%
\hfill
\begin{subfigure}{.32\textwidth}
  \centering
  \includegraphics[width=\linewidth]{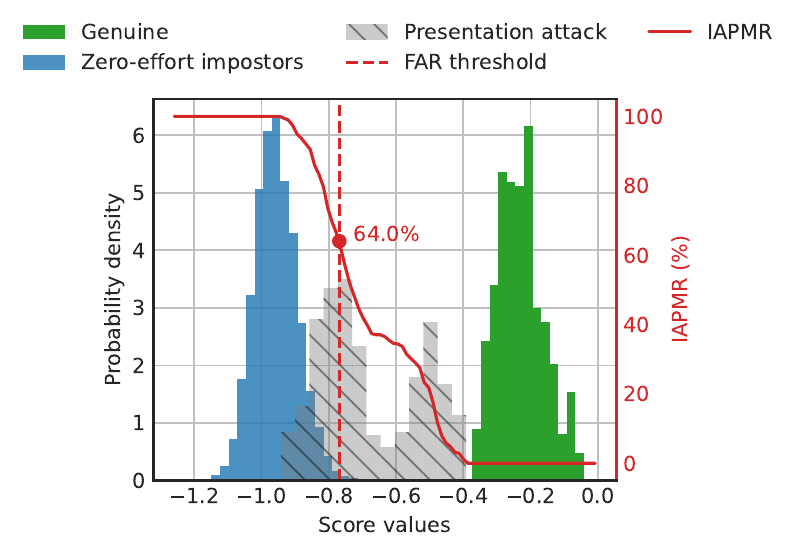}
  \caption{DIU - Masks}
\end{subfigure}

\begin{subfigure}{.32\textwidth}
  \centering
  \includegraphics[width=\linewidth]{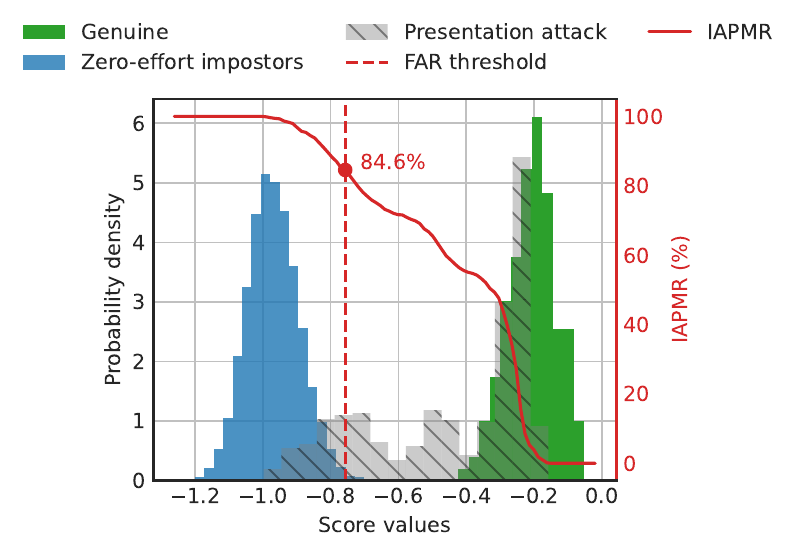}
  \caption{SSMB - All PAs}
\end{subfigure}%
\hfill
\begin{subfigure}{.32\textwidth}
  \centering
  \includegraphics[width=\linewidth]{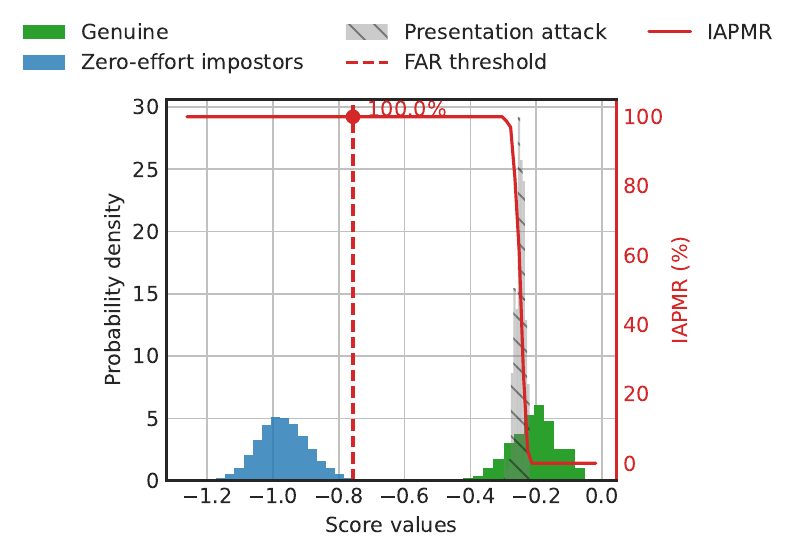}
  \caption{SSMB - Laser Photos}
\end{subfigure}%
\hfill
\begin{subfigure}{.32\textwidth}
  \centering
  \includegraphics[width=\linewidth]{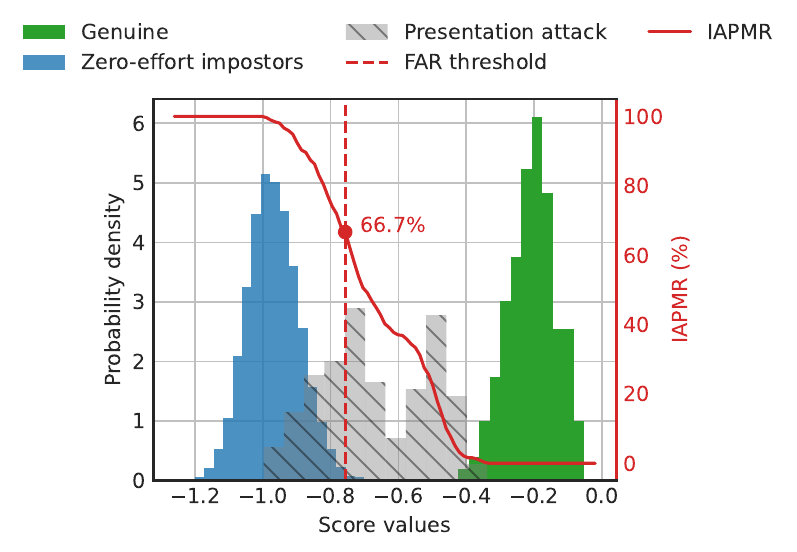}
  \caption{SSMB - Masks}
\end{subfigure}

\caption{Score distributions (VIS-VIS Protocol) for two HFR systems (first row DIU \cite{george2024heterogeneous}, second row SSMB \cite{george2024modality}) with different PA combinations (All PAs, Laser Photos, Masks).
Each plot shows histograms of genuine (green), ZEI (blue), and attack (gray) scores. The red dashed line marks the FMR 0.1\% threshold (from the licit protocol's development group), while the solid red curve represents IAPMR across thresholds. The IAPMR at the given threshold is found at the curve's intersection with the dashed line.}

\label{fig:vulnhist_vis_rgb}
\end{figure*}

To further analyze system vulnerability, we plot the similarity score distributions for each of the potent attack species under both VIS-NIR (Fig. \ref{fig:vulnhist_vis_nir}) and VIS-VIS (Fig. \ref{fig:vulnhist_vis_rgb}) settings. These plots compare the scores of successful attacks against genuine comparison scores and zero-effort impostor (ZEI) scores. In the VIS-NIR protocol, the distribution of scores for laser photo attacks closely overlaps with that of genuine comparisons, highlighting the CFR system's significant vulnerability to this particular attack.

\textbf{COTS-PAD evaluation}: Experiments in the previous sections have demonstrated that unprotected cross-spectral and heterogeneous face recognition (CFR/HFR) systems are highly vulnerable to specific categories of presentation attacks. To assess the feasibility of enhancing the security of such systems, we evaluate the performance of a presentation attack detection (PAD) from the same commercial off-the-shelf (COTS) face recognition system (COTS-PAD). This evaluation is conducted on the probe samples of the evaluation set, which is the only subset containing presentation attack samples. Given that only the evaluation set includes attacks, we determine the equal error rate (EER) threshold directly on this set. Using this threshold, we compute the Attack Presentation Classification Error Rate (APCER) across various attack types. In addition, we report the corresponding Bonafide Presentation Classification Error Rate (BPCER) and the Average Classification Error Rate (ACER). To facilitate a comparative analysis of PAD effectiveness under different spectral conditions, we perform this evaluation for probe samples in both VIS-VIS and VIS-NIR protocols. Preliminary results, presented in Table \ref{tab:cots_pad}, show that PAD performance in the NIR domain is particularly limited. Notably, laser-printed photo attacks are the most difficult to detect in this domain, with an APCER of 98.2\%. These findings underscore the urgent need for more robust and specialized PAD solutions to effectively secure CFR systems.

\begin{table}[ht]
\caption{PAD Metrics for COTS-PAD system across VIS and NIR images}

\centering
  \resizebox{0.95\columnwidth}{!}{%

\begin{tabular}{l|c|c}
\toprule
\textbf{Metric} & \textbf{VIS (COTS-PAD)} & \textbf{NIR (COTS-PAD)} \\
\midrule
APCER (flexiblemask) & 39.0\% & 81.9\% \\
APCER (rigidmask) & \textbf{39.6}\% & 56.0\% \\
APCER (prints) & 37.7\% & 0.0\% \\
APCER (laser\_photo) & 21.8\% & \textbf{98.2}\% \\
APCER (replay) & 26.9\% & 0.0\% \\
APCER (papermask) & 24.4\% & 35.6\% \\ \midrule
APCER (AP) & 39.6\% & 98.2\% \\
BPCER & 32.1\% & 29.8\% \\
ACER & 35.9\% & 64.0\% \\
\bottomrule
\end{tabular}
}
\label{tab:cots_pad}
\end{table}

\subsection{Discussion}

Contrary to earlier assumptions, our experimental results clearly demonstrate that CFR systems remain vulnerable to presentation attacks. Although the effectiveness of many common attack types is reduced in the VIS-NIR setting, printed photos produced using a laser printer emerged as the most successful attack method, achieving IAPMR values of 96.97\%, 100\% and 100\% across the evaluated systems. These attacks are both simple to produce and easy to carry out, highlighting a critical vulnerability in current CFR approaches. 

The evaluation with COTS-PAD showed that the detection of attacks in NIR is specifically hard. This finding emphasizes the need for dedicated Presentation Attack Detection (PAD) systems tailored for CFR scenarios. While previous studies have shown that combining RGB and NIR modalities can enhance PAD performance, CFR use cases typically lack access to RGB data during the probe phase. As a result, there is a pressing requirement to develop specialized PAD systems capable of operating solely on NIR modality. Furthermore, the high success rate of laser-printed photo attacks suggests that NIR-reflective inks can be exploited to create even more potent attacks against VIS-NIR CFR systems. It is also important to note that this vulnerability analysis of CFR focuses solely on impersonation attacks. However, the design and execution of obfuscation attacks, particularly in the NIR domain may be significantly easier due to the availability of reflective or absorptive inks. This presents an additional and critical challenge for the development of effective PAD schemes.

\section{Conclusion}

In this work, we have demonstrated that cross-spectral face recognition (CFR) systems remain highly vulnerable to certain types of presentation attacks. Most notably, simple attacks using laser-printed photos were found to be particularly effective, achieving very high IAPMR values of 96.97\%, 100\%, and 100\% across the evaluated systems. These results challenge the assumption that CFR systems inherently offer improved security over traditional face recognition systems. While many common attack types exhibit reduced effectiveness in the VIS-NIR setting due to spectral mismatch and reflectance differences, the success of laser-printed photo attacks highlights a significant blind spot. A key limitation of most of the existing presentation attack detection (PAD) methods is their dependence on RGB data, which is often unavailable in the probe phase of CFR pipelines. Moreover, the potential use of NIR-reflective or absorptive inks introduces a new and underexplored class of spoofing materials that could further compromise system integrity. While our study focuses solely on impersonation attacks, obfuscation attacks in the NIR domain may be even easier to execute, posing an additional challenge for PAD.
These results highlights the need for PAD methods specifically designed for NIR-only scenarios. Future work will focus on developing robust NIR-based PAD solutions, with an emphasis on detecting these spoofing artifacts.

\section{Acknowledgments}

This research was funded by the European Union project CarMen (Grant Agreement No. 101168325) as well as the Swiss Center for Biometrics Research and Testing.

{\small
\bibliographystyle{ieee}
\bibliography{egbib}
}

\end{document}